%% file: tagCNN.tex
\documentclass[11pt]{article}
\usepackage{coling2014}
\usepackage{times}
\usepackage{url}
\usepackage{latexsym}
\usepackage{graphicx}
\usepackage{multirow}
\usepackage{amsmath}
\input{packages}
\input{MACPdef}
\input{MACPdef-ams}

\title{Encoding Source Language with Convolutional Neural Network \\ for Machine Translation}

\author{Fandong Meng$^1$ \ Zhengdong Lu$^2$ \ Mingxuan Wang$^1$ 
	     Hang Li$^2$ \ Wenbin Jiang$^1$ \ Qun Liu$^{3,1}$ \\
        $^1$Institute of Computing Technology, Chinese Academy of Sciences\\
        {\tt \{mengfandong,wangmingxuan,jiangwenbin,liuqun\}@ict.ac.cn}\\
        $^2$Noah's Ark Lab, Huawei Technologies\\
        {\tt \{Lu.Zhengdong,HangLi.HL\}@huawei.com}\\
        $^3$ADAPT Centre, School of Computing, Dublin City University\\
}

\date{}

\begin{document}
\maketitle
\begin{abstract}
The recently proposed neural network joint model (NNJM)~\cite{devlin2014} augments the n-gram target language model with a heuristically chosen source context window, achieving state-of-the-art performance in SMT. In this paper, we give a more systematic treatment by summarizing the relevant source information through a convolutional architecture guided by the target information. With different guiding signals during decoding, our specifically designed convolution+gating architectures can pinpoint the parts of a source sentence that are relevant to predicting a target word, and fuse them with the context of entire source sentence to form a unified representation. This representation, together with target language words, are fed to a deep neural network (DNN) to form a stronger NNJM. Experiments on two NIST Chinese-English translation tasks show that the proposed model can achieve significant improvements over the previous NNJM by up to +1.08 BLEU points on average.
\end{abstract}

\section{Introduction} 
Learning of continuous space representation for source language has attracted much attention in both traditional statistical machine translation (SMT) and neural machine translation (NMT). Various models, mostly neural network-based, have been proposed for representing the source sentence, mainly as the encoder part in an encoder-decoder framework ~\cite{Bengio03aneural,auli2013,kalchbrenner2013,cho2014,googleS2S}. There has been some quite recent work on encoding only ``relevant" part of source sentence during the decoding process, most notably neural network joint model (NNJM) in \cite{devlin2014}, which extends the $n$-grams target language model by additionally taking a fixed-length window of source sentence, achieving state-of-the-art performance in statistical machine translation. 

In this paper, we propose novel convolutional architectures to dynamically encode the relevant information in the source language. Our model covers the entire source sentence, but can effectively find and properly summarize the relevant parts, guided by the information from the target language. With the guiding signals during decoding, our specifically designed convolution architectures can pinpoint the parts of a source sentence that are relevant to predicting a target word, and fuse them with the context of entire source sentence to form a unified representation. This representation, together with target words, are fed to a deep neural network (DNN) to form a stronger NNJM. Since our proposed joint model is purely lexicalized, it can be integrated into any SMT decoder as a feature.

Two variants of the joint model are also proposed, with coined name $tag$CNN and $in$CNN, with different guiding signals used from the decoding process. We integrate the proposed joint models into a state-of-the-art dependency-to-string translation system~\cite{xie2011novel} to evaluate their effectiveness. Experiments on NIST Chinese-English translation tasks show that our model is able to achieve significant improvements of +2.0 BLEU points on average over the baseline. Our model also outperforms~\newcite{devlin2014}'s NNJM by up to +1.08 BLEU points. 

\begin{figure*}[t!]
\begin{center}
      \includegraphics[width=0.40\textwidth]{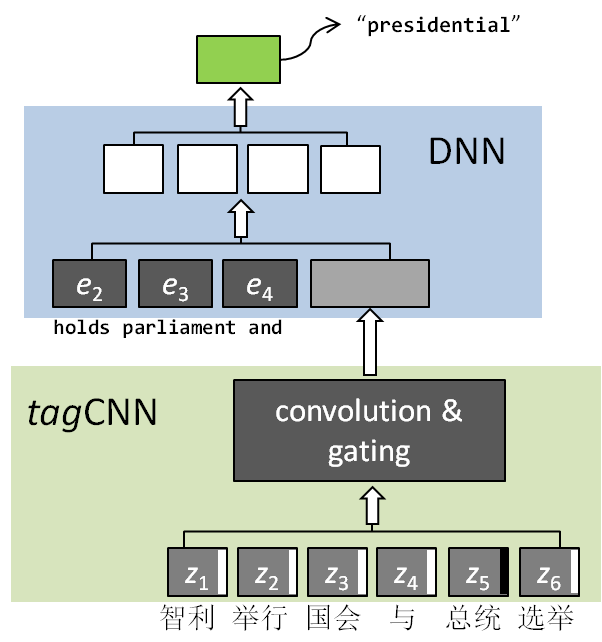}
      \hspace{30pt}
      \includegraphics[width=0.45\textwidth]{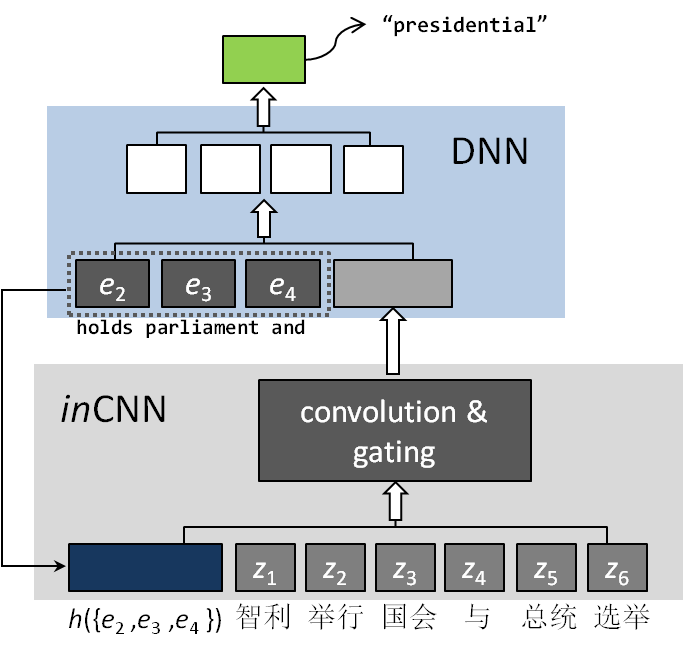}\\
      (a) $tag$CNN  \hspace{150pt} (b) $in$CNN
    \caption{Illustration for joint LM based on CNN encoder.}
    \label{f:JLM}
  \end{center}
\end{figure*}

\paragraph{RoadMap:}
In the remainder of this paper, we start with a brief overview of joint language model in Section~\ref{s:JLM}, while the convolutional encoders, as the key component of which, will be described in detail in Section~\ref{s:cnn}. Then in Section \ref{s:decoder} we discuss the decoding algorithm with the proposed models. The experiment results are reported in Section~\ref{s:expts}, followed by Section~\ref{s:related} and~\ref{s:conclusion} for related work  and conclusion.

\section{Joint Language Model} \label{s:JLM}
Our joint model with CNN encoders can be illustrated in Figure \ref{f:JLM} (a) \& (b), which consists 1) a CNN encoder, namely $tag$CNN or $in$CNN, to represent the information in the source sentences, and 2) an NN-based model for predicting the next words, with representations from CNN encoders and the history words in target sentence as inputs.

In the joint language model,  the probability of the target word $\mathbf{e}_n$,  given previous $k$ target words $\{\mathbf{e}_{n-k}, \cdots\hspace{-3pt}, \mathbf{e}_{n-1}\}$ and the representations from CNN-encoders for source sentence $S$ are
{
\setlength\abovedisplayskip{4pt}
\setlength\belowdisplayskip{4pt}
\begin{eqnarray*}
\hspace{-2pt}&\text{$tag$CNN:}&p(\mathbf{e}_n |\phi_1(S, \{a(\mathbf{e}_n)\}), \, \{\mathbf{e}\}_{n-k}^{n-1})\\
\hspace{-2pt}&\text{$in$CNN:}&p(\mathbf{e}_n |\,\phi_2(S,  h(\{\mathbf{e}\}_{n-k}^{n-1})), \, \{\mathbf{e}\}_{n-k}^{n-1}),
\end{eqnarray*}
}where $\phi_1(S, \{a(\mathbf{e}_n)\})$ stands for the representation given by $tag$CNN with the set of indexes $\{a(\mathbf{e}_n)\}$ of source words aligned to the target word $\mathbf{e}_n$, and $\phi_2(S,  h(\{\mathbf{e}\}_{n-k}^{n-1}))$ stands for the representation from $in$CNN with the attention signal $ h(\{\mathbf{e}\}_{n-k}^{n-1})$.

Let us use the example in Figure \ref{f:JLM}, where the task is to translate the Chinese sentence
\begin{figure}[h!]
\begin{center}
      \includegraphics[width=0.50\textwidth]{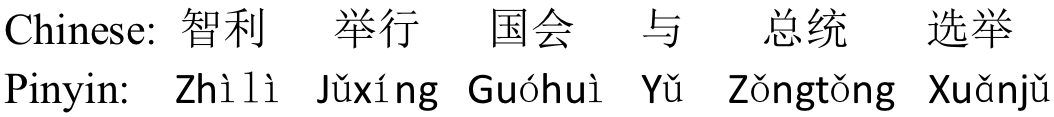}
  \end{center} \vspace{-10pt}
\end{figure}

\noindent into English. In evaluating a target language sequence ``\texttt{holds parliament and presidential}", with  ``\texttt{holds parliament and}" as the proceeding words (assume 4-gram LM), and the affiliated source word\footnote{For an aligned target word, we take its aligned source words as its affiliated source words. And for an unaligned word, we inherit its affiliation from the closest aligned word, with preference given to the right~\cite{devlin2014}. Since the word alignment is of many-to-many, one target word may has multi affiliated source words.}  of ``\texttt{presidential}" being ``\texttt{Z\v{o}ngt\v{o}ng}" (determined by word alignment), $tag$CNN generates $\phi_1(S, \text{\{4\}})$ (the index of ``\texttt{Z\v{o}ngt\v{o}ng}" is 4), and $in$CNN generates $\phi_2(S,  h(\texttt{holds parliament and}))$. The DNN component then takes \texttt{"holds parliament and"} and ($\phi_1$ or $\phi_2$) as input to give the conditional probability for next word, e.g., $p(\texttt{"presidential"}|\phi_{1|2}, \, \texttt{\{holds}, \, \texttt{parliament}, \, \texttt{and\}})$.

\section{Convolutional Models} \label{s:cnn}
We start with the  generic architecture for convolutional encoder, and then proceed to $tag$CNN and $in$CNN as two extensions.

\subsection{Generic CNN Encoder} \label{s:genericCNN}
The basic architecture is of a generic CNN encoder is illustrated in Figure \ref{f:convs} (a), which has a fixed architecture consisting of six layers:
\begin{description}
  \item[Layer-0:] the input layer, which takes words in the form of embedding vectors. In our work, we set the maximum length of sentences to 40 words. For sentences shorter than that, we put zero padding at the beginning of sentences.  
  \item[Layer-1:] a convolution layer after Layer-0, with window size = 3. As will be discussed in Section \ref{s:tagCNN} and \ref{s:inCNN}, the guiding signal are injected into this layer for ``guided version". 
  \item[Layer-2:] a local gating layer after Layer-1, which simply takes a weighted sum over feature-maps in non-adjacent window with size = 2. 
  \item[Layer-3:] a convolution layer after Layer-2, we perform another convolution with window size = 3.  
  \item[Layer-4:] we perform a global gating over feature-maps on Layer-3.  
  \item[Layer-5:] fully connected weights that maps the output of Layer-4 to this layer as the final representation.
\end{description}

\subsubsection{Convolution} 
\begin{figure*}[t!]
\begin{center}
      \includegraphics[width=0.95\textwidth]{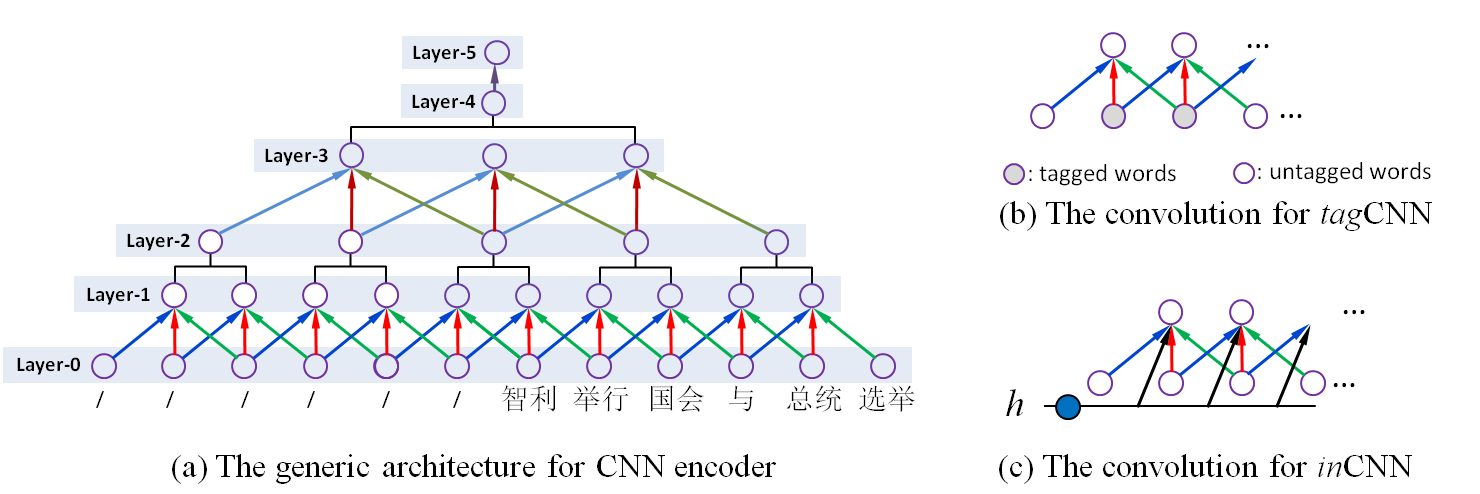} \hspace{30pt}
      \vspace{-10pt}
    \caption{Illustration for the CNN encoders.}
    \label{f:convs}
  \end{center} 
\end{figure*}

As shown in Figure \ref{f:convs} (a), the convolution in Layer-1 operates on sliding windows of words (width $k_1$), and the similar definition of windows carries over to higher layers. Formally, for source sentence input $\mathbf{x} \hspace{-3pt}=\hspace{-3pt} \{\x_1,\cdots, \x_N\}$, the convolution unit for feature map of type-$f$ (among $F_\ell$ of them)  on Layer-$\ell$ is 
\begin{multline}
z^{(\ell,f)}_{i}(\mathbf{x}) =  \sigma(\w^{(\ell,f)} \hat{\mathbf{z}}^{(\ell-1)}_{i} + b^{(\ell,f)}), \hspace{10pt} \ell = 1,3, \hspace{10pt} f = 1,2,\cdots,F_{\ell}
\end{multline}
where
\begin{itemize}
  \item $z^{(\ell,f)}_{i}(\mathbf{x})$ gives the output of feature map of type-$f$  for location $i$ in Layer-$\ell$;
   \item $\w^{(\ell, f)}$ is the parameters for $f$ on Layer-$\ell$;
   \item $\sigma(\cdot)$ is the Sigmoid activation function;
  \item $\hat{\mathbf{z}}^{(\ell-1)}_{i}$ denotes the segment of Layer-$\ell\hspace{-3pt}-\hspace{-3pt}1$ for the convolution at location $i$ , while
      \begin{equation*}
      \hat{\mathbf{z}}^{(0)}_{i} \overset{\text{def}}{=}  [ \x_{i}^\top, \;\, \x_{i+1}^\top,\;\, \x_{i+2}^\top]^\top
      \end{equation*}
      concatenates the vectors for 3 words from sentence input $\mathbf{x}$.
\end{itemize}

\subsubsection{Gating}
\label{s:gate}
Previous CNNs, including those for NLP tasks~\cite{hu2014convolutional,KalchbrennerACL2014}, take a straightforward convolution-pooling strategy, in which the ``fusion" decisions (e.g., selecting the largest one in max-pooling) are based on the values of feature-maps. This is essentially a soft template matching, which works for tasks like classification, but harmful for keeping the composition functionality of convolution, which is critical for modeling sentences. In this paper, we propose to use separate gating unit to release the score function duty from the convolution, and let it focus on composition.

We take two types of gating: 1) for Layer-2, we take a local gating with non-overlapping windows (size = 2) on the feature-maps of convolutional Layer-1 for representation of segments, and 2) for Layer-4, we take a global gating to fuse all the segments for a global representation. We found that this gating strategy can considerably improve the performance of both $tag$CNN and $in$CNN over pooling.

\begin{itemize}
  \item {\bf Local Gating:} On Layer-1, for every gating window, we first find its original input (before convolution) on Layer-0, and merge them for the input of the gating network. For example, for the two windows: word (3,4,5) and word (4,5,6) on Layer-0, we use concatenated vector consisting of embedding for word (3,4,5,6) as the input of the local gating network (a logistic regression model) to determine the weight for the convolution result of the two windows (on Layer-1), and the weighted sum are the output of Layer-2. 
  \item {\bf Global Gating:} On Layer-3, for feature-maps at each location $i$, denoted $\z_i^{(3)}$, the global gating network (essentially soft-max, parameterized $\w_{g}$), assigns a normalized weight 
\[
\omega(\z_i^{(3)}) = e^{\w_{g}^\top \z_i^{(3)}}/\sum_j e^{\w_{g}^\top \z_j^{(3)}}, \vspace{-10pt}
\]
and the gated representation on Layer-4 is given by the weighted sum $\sum_i \omega(\z_i^{(3)}) \z_i^{(3)}$.
\end{itemize}

\subsubsection{Training of CNN encoders}
The CNN encoders, including $tag$CNN and $in$CNN that will be discussed right below, are trained in a joint language model described in Section~\ref{s:JLM}, along with the following parameters 
\begin{itemize}
  \item the embedding of the words on source and the proceeding words on target; 
  \item the parameters for the DNN of joint language model, include the parameters of soft-max for word probability. 
\end{itemize}

The training procedure is identical to that of neural network language model, except that the parallel corpus is used instead of a monolingual corpus. We seek to maximize the log-likelihood of training samples, with one sample for every target word in the parallel corpus. Optimization is performed with the conventional back-propagation, implemented as stochastic gradient descent~\cite{lecun-98b} with mini-batches.

\subsection{$tag$CNN} \label{s:tagCNN}
$tag$CNN inherits the convolution and gating from generic CNN (as described in Section \ref{s:genericCNN}), with the only modification in the input layer.
As shown in Figure \ref{f:convs} (b), in $tag$CNN, we append an extra tagging bit ($0$ or $1$) to the embedding of words in the input layer to indicate whether it is one of affiliated words
\[
\x_i^{(\textsc{Aff})} = [\x_i^\top\;1]^\top,\;\;\; \x_j^{(\textsc{non-Aff})} = [\x_j^\top\;0]^\top. 
\]
Those extended word embedding will then be treated as regular word-embedding in the convolutional neural network. This particular encoding strategy can be extended to embed more complicated dependency relation in source language, as will be described in Section~\ref{s:dep}.

This particular ``tag" will be activated in a parameterized way during the training for predicting the target words. In other words, the supervised signal from the words to predict will find, through layers of back-propagation, the importance of the tag bit in the ``affiliated words" in the source language, and learn to put proper weight on it to make tagged words stand out and adjust other parameters in $tag$CNN accordingly for the optimal predictive performance. In doing so, the joint model can pinpoint the parts of a source sentence that are relevant to predicting a target word through the already learned word alignment.

\subsection{$in$CNN} \label{s:inCNN}
Unlike $tag$CNN, which directly tells the location of affiliated words to the CNN encoder, $in$CNN sends the information about the proceeding words in target side to the convolutional encoder to help retrieve the information relevant for predicting the next word. This is essentially a particular case of attention model, analogous to the automatic alignment mechanism in~\cite{bahdanau2014neural}, where the attention signal is from the state of a generative recurrent neural network (RNN) as decoder.

Basically, the information from proceeding words, denoted as $h(\{\mathbf{e}\}_{n-k}^{n-1})$, is injected into every convolution window in the source language sentence, as illustrated in Figure \ref{f:convs} (c). More specifically, for the window indexed by $t$, the input to convolution is given by the concatenated vector 
\[
\hat{\z}_t = [h(\{\mathbf{e}\}_{n-k}^{n-1}),\; \x_{t}^\top, \;\x_{t+1}^\top, \;\x_{t+2}^\top]^\top. 
\]
In this work, we use a DNN to transform the vector concatenated from word-embedding for words $\{\mathbf{e}_{n-k}\cdots, \,\mathbf{e}_{n-k}\}$ into $h(\{\mathbf{e}\}_{n-k}^{n-1})$, with sigmoid activation function. Through layers of convolution and gating, $in$CNN can 1) retrieve the relevant segments of source sentences, and 2) compose and transform the retrieved segments into representation recognizable by the DNN in predicting the words in target language. Different from that of $tag$CNN, $in$CNN uses information from proceeding words, hence provides complementary information in the augmented joint language model of $tag$CNN. This has been empirically verified when using feature based on $tag$CNN and that based on $in$CNN in decoding with greater improvement.

\begin{figure*}[t!]
\begin{center}
      \includegraphics[width=0.9\textwidth]{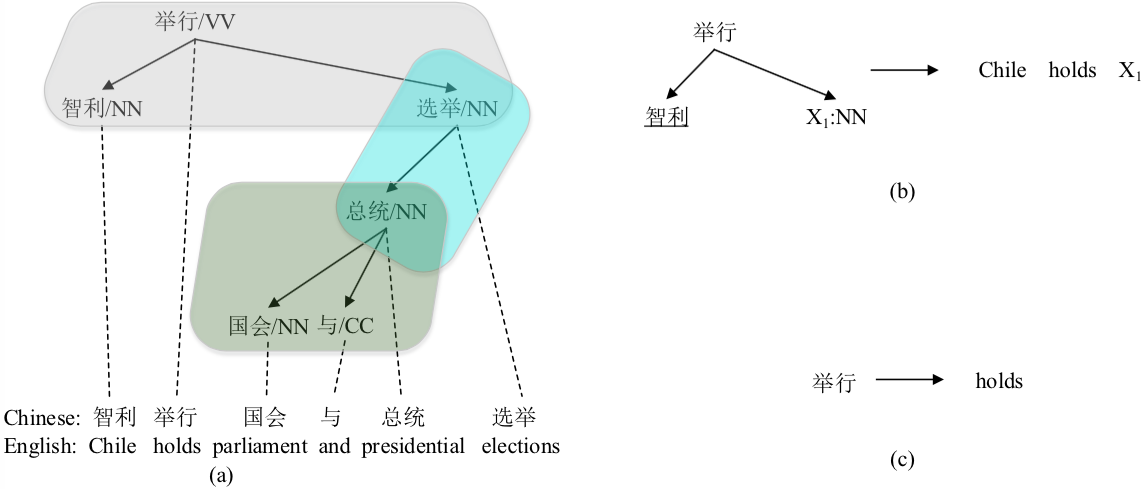}
    \caption{Illustration for a dependency tree (a) with three head-dependents relations in shadow, an example of head-dependents relation rule (b) for the top level of (a), and an example of head rule (c). ``$X_1$:NN" indicates a substitution site that can be replaced by a subtree whose root has part-of-speech ``NN". The underline denotes a leaf node.}
    \label{f:deptree}
  \end{center}
\end{figure*}

\section{Decoding with the Joint Model} \label{s:decoder}
Our joint model is purely lexicalized, and therefore can be integrated into \emph{any} SMT decoders as a feature. For a hierarchical SMT decoder, we adopt the integrating method proposed by~\newcite{devlin2014}. As inherited from the $n$-gram language model for performing hierarchical decoding, the leftmost and rightmost $n\hspace{-1pt}-\hspace{-1pt}1$ words from each constituent should be stored in the state space. We extend the state space to also include the indexes of the affiliated source words for each of these edge words. For an aligned target word, we take its aligned source words as its affiliated source words. And for an unaligned word, we use the affiliation heuristic adopted by~\newcite{devlin2014}. In this paper, we integrate the joint model into the state-of-the-art dependency-to-string machine translation decoder as a case study to test the efficacy of our proposed approaches. We will briefly describe the dependency-to-string translation model and then the description of MT system. 

\subsection{Dependency-to-String Translation} 
In  this paper, we use a state-of-the-art dependency-to-string~\cite{xie2011novel} decoder (Dep2Str), which is also a hierarchical decoder. This dependency-to-string model employs rules that represent the source side as head-dependents relations and the target side as strings. A head-dependents relation (HDR) is composed of a head and all its dependents in dependency trees.  Figure~\ref{f:deptree} shows a dependency tree (a) with three HDRs (in shadow), an example of HDR rule (b) for the top level of (a), and an example of head rule (c). HDR rules are constructed from head-dependents relations. HDR rules can act as both translation rules and reordering rules. And head rules are used for translating source words.

We adopt the decoder proposed by~\newcite{meng2013} as a variant of Dep2Str translation that is easier to implement with comparable performance. Basically they extract the HDR rules with GHKM~\cite{galley2004s} algorithm. For the decoding procedure, given a source dependency tree $T$, the decoder transverses $T$ in post-order. The bottom-up chart-based decoding algorithm with cube pruning~\cite{chiang2007hierarchical,huang2007forest} is used to find the $k$-best items for each node.

\subsection{MT Decoder}
Following~\newcite{och2002discriminative}, we use a general log\-linear framework. Let $d$ be a derivation that convert a source dependency tree into a target string $e$. The probability of $d$ is defined as: 
\begin{equation}
P(d) \propto \prod_{i} \phi_i(d)^{\lambda_i}
\end{equation}
where $\phi_i$ are features defined on derivations and $\lambda_i$ are the corresponding weights. Our decoder contains the following features:\\
{\bf Baseline Features:}
\begin{itemize}
\item translation probabilities $P(t|s)$ and $P(s|t)$ of HDR rules; 
\item lexical translation probabilities $P_{\textsc{lex}}(t|s)$ and $P_{\textsc{lex}}(s|t)$ of HDR rules; 
\item rule penalty $\exp(-1)$; 
\item pseudo translation rule penalty $\exp(-1)$; 
\item target word penalty $\exp(|e|)$; 
\item $n$-gram language model $P_{\textsc{lm}}(e)$; 
\end{itemize}
{\bf Proposed Features:} 
\begin{itemize}
\item $n$-gram $tag$CNN joint language model $P_{\textsc{tlm}}(e)$; 
\item $n$-gram $in$CNN joint language model $P_{\textsc{ilm}}(e)$. 
\end{itemize}
Our baseline decoder contains the first eight features. The pseudo translation rule (constructed according to the word order of a HDR) is to ensure the complete translation when no matched rules is found during decoding. The weights of all these features are tuned via minimum error rate training (MERT)~\cite{och2003minimum}. For the dependency-to-string decoder, we set rule-threshold and stack-threshold to $10^{-3}$, rule-limit to 100, stack-limit to 200.

\section{Experiments} \label{s:expts}
The experiments in this Section are designed to answer the following questions:  
\begin{enumerate}
  \item Are our $tag$CNN and $in$CNN joint language models able to improve translation quality, and are they complementary to each other?  
  \item Do $in$CNN and $tag$CNN benefit from their guiding signal, compared to a generic CNN? 
  \item For $tag$CNN, is it helpful to embed more dependency structure,  e.g., dependency head of each affiliated word, as additional information? 
  \item Can our gating strategy improve the performance over max-pooling?
\end{enumerate}

\subsection{Setup} \label{s:setup}
\paragraph{Data:} Our training data are extracted from LDC  data\footnote{The corpora include LDC2002E18, LDC2003E07, LDC2003E14, LDC2004T07, LDC2005T06.}. We only keep the sentence pairs that the length of source part no longer than 40 words, which covers over 90\% of the sentence. The bilingual training data consist of 221K sentence pairs, containing 5.0 million Chinese words and 6.8 million English words. The development set is NIST MT03 (795 sentences)  and test sets are MT04 (1499 sentences) and MT05 (917 sentences) after filtering with length limit. 

\paragraph{Preprocessing:} The word alignments are obtained with GIZA++~\cite{och2003systematic} on the corpora in both directions, using the ``grow-diag-final-and" balance strategy~\cite{koehn2003statistical}. We adopt SRI Language Modeling Toolkit~\cite{stolcke2002srilm} to train a 4-gram language model with modified Kneser-Ney smoothing on the Xinhua portion of the English Gigaword corpus (306 million words). We parse the Chinese sentences with Stanford Parser into projective dependency trees. 

\paragraph{Optimization of NN:} In training the neural network, we limit the source and target vocabulary to the most frequent 20K words for both Chinese and English, covering approximately 97\% and 99\% of two corpus respectively. All the out-of-vocabulary words are mapped to a special token \texttt{\small UNK}.  We used stochastic gradient descent to train the joint model, setting the size of minibatch to 500. All joint models used a 3-word target history (i.e., 4-gram LM). The dimension of word embedding and the attention signal $h(\{\mathbf{e}\}_{n-k}^{n-1})$ for $in$CNN are 100. For the convolution layers (Layer 1 and Layer 3), we apply 100 filters. And the final representation of CNN encoders is a vector with dimension 100. The final DNN layer of our joint model is the standard multi-layer perceptron with softmax at the top layer. 

\paragraph{Metric:} We use the case-insensitive 4-gram NIST BLEU\footnote{\url{ftp://jaguar.ncsl.nist.gov/mt/resources/mteval-v11b.pl}} as our evaluation metric, with statistical significance test with \emph{sign-test}~\cite{collins2005clause} between the proposed models and two baselines.

\subsection{Setting for Model Comparisons}
We use the $tag$CNN and $in$CNN joint language models as additional decoding features to a dependency-to-string baseline system (Dep2Str), and compare them to the neural network joint model with 11 source context words~\cite{devlin2014}. We use the implementation of an open source toolkit\footnote{http://nlg.isi.edu/software/nplm/} with default configuration except the global settings described in Section~\ref{s:setup}. Since our $tag$CNN and $in$CNN models are source-to-target and left-to-right (on target side), we only take the source-to-target and left-to-right type NNJM in~\cite{devlin2014} in comparison. We call this type NNJM as BBN-JM hereafter. Although the BBN-JM in~\cite{devlin2014} is originally tested in the hierarchical phrase-based~\cite{chiang2007hierarchical}  SMT and string-to-dependency~\cite{shen2008new} SMT, it is fairly versatile and can be readily integrated into Dep2Str.

\subsection{The Main Results }
The main results of different models are given in Table~\ref{t:model-cpm}. Before proceeding to more detailed comparison, we first observe that 
\begin{itemize}
  \item the baseline Dep2Str system gives BLEU 0.5+ higher than the open-source phrase-based system Moses~\cite{koehn2007};
  \item BBN-JM can give about +0.92 BLEU score over Dep2Str, a result similar as reported in~\cite{devlin2014}.  
\end{itemize}

Clearly from Table~\ref{t:model-cpm}, $tag$CNN and $in$CNN improve upon the Dep2Str baseline by +1.28 and +1.75 BLEU, outperforming BBN-JM in the same setting by respectively +0.36 and +0.83 BLEU,  averaged on NIST MT04 and MT05. These indicate that $tag$CNN and $in$CNN can individually provide discriminative information in decoding. It is worth noting that $in$CNN appears to be more informative than the affiliated words suggested by the word alignment (GIZA++). We conjecture that this is due to the following two facts
\begin{itemize} 
  \item $in$CNN avoids the propagation of mistakes and artifacts in the already learned word alignment;
  \item the guiding signal in $in$CNN provides complementary information to evaluate the translation.
\end{itemize}
 Moreover, when $tag$CNN and $in$CNN are both used in decoding, it can further increase its winning margin over BBN-JM to +1.08 BLEU points (in the last row of Table~\ref{t:model-cpm}), indicating that the two models with different guiding signals are complementary to each other.

\begin{table*}[t!]
\begin{center}
\scalebox{0.9}{
\begin{tabular}{l|lll}
\hline
\textbf{Systems} & \textbf{MT04} & \textbf{MT05} & \textbf{Average} \\
\hline\hline
Moses          	& 34.33      & 31.75         & 33.04 \\
Dep2Str 		& 34.89      & 32.24         & 33.57 \\
\hline
\hspace{40pt}$+$ BBN-JM~\cite{devlin2014}          & 36.11            & 32.86              & 34.49 \\
\hline
\hspace{40pt}$+$ CNN (generic)  			      & 36.12*           & 33.07*             & 34.60 \\
\hspace{40pt}$+$ $tag$CNN       			      & 36.33*           & {\bf 33.37}*    & 34.85 \\
\hspace{40pt}$+$ $in$CNN        			      & {\bf 36.92}*   & {\bf 33.72}*    & 35.32 \\
\hspace{40pt}$+$ $tag$CNN $+$ $in$CNN   	      & {\bf 36.94}*   & {\bf 34.20}*    & 35.57 \\
\hline
\end{tabular}
}
\end{center} 
\caption{\label{t:model-cpm} BLEU-4 scores (\%) on NIST MT04-test and MT05-test, of Moses (default settings), dependency-to-string baseline system (Dep2Str), and different features on top of Dep2Str: neural network joint model (BBN-JM),  generic CNN, $tag$CNN, $in$CNN and the combination of $tag$CNN and $in$CNN. The boldface numbers and superscript $^*$ indicate that the results are significantly better (p$<$0.01) than those of the BBN-JM and the Dep2Str baseline respectively. ``$+$" stands for adding the corresponding feature to Dep2Str.}
\end{table*}

\paragraph{The Role of Guiding Signal}
It is slight surprising that the generic CNN can also achieve the gain on BLEU similar to that of BBN-JM, since intuitively generic CNN encodes the entire sentence and the representations should in general far from optimal representation for joint language model. The reason, as we conjecture, is CNN yields fairly informative summarization of the sentence (thanks to its sophisticated convolution and gating architecture), which makes up some of its loss on resolution and relevant parts of the source senescence. That said, the guiding signal in both $tag$CNN and $in$CNN are crucial to the power of CNN-based encoder, as can be easily seen from the difference between the BLEU scores achieved by generic CNN, $tag$CNN, and $in$CNN. Indeed, with the signal from the already learned word alignment, $tag$CNN can gain +0.25 BLEU over its generic counterpart, while for $in$CNN with the guiding signal from the proceeding words in target,  the gain is more saliently +0.72 BLEU.

\subsection{Dependency Head in $tag$CNN} \label{s:dep}
In this section, we study whether $tag$CNN can further benefit from encoding richer dependency structure in source language in the input. More specifically, the dependency head words can be used to further improve $tag$CNN model. As described in Section~\ref{s:tagCNN}, in $tag$CNN, we append a tagging bit (0 or 1) to the embedding of words in the input layer as tags on whether they are affiliated source words. To incorporate dependency head information, we extend the tagging rule in Section~\ref{s:tagCNN} to add another tagging bit (0 or 1) to the word-embedding for original $tag$CNN to indicate whether it is part of dependency heads of the affiliated words. For example, if $\mathbf{x}_i$ is the embedding of an affiliated source word and $\mathbf{x}_j$ the dependency head of word $\mathbf{x}_i$, the extended input of tagCNN would contain
\begin{eqnarray*}
\x_i^{(\textsc{Aff},\;\textsc{non-Head})} &=& [\x_i^\top\;\;1\;\;0]^\top\\
\x_j^{(\textsc{non-Aff},\;\textsc{Head})} &=& [\x_j^\top\;\;0\;\;1]^\top
\end{eqnarray*}
If the affiliated source word is the root of a sentence, we only append 0 as the second tagging bit since the root has no dependency head. From Table~\ref{t:deptag-cpm}, with the help of dependency head information, we can improve $tag$CNN by +0.23 BLEU points averagely on two test sets.

\begin{table}[t]
\begin{center}
\scalebox{0.9}{
\begin{tabular}{l|ccc}
\hline
\textbf{Systems} & \textbf{MT04} & \textbf{MT05} & \textbf{Average} \\
\hline\hline
Dep2str &  34.89 & 32.24 & 33.57 \\
\hline
+$tag$CNN & 36.33 & 33.37 & 34.85 \\
+$tag$CNN\_dep & 36.54 & 33.61 & 35.08 \\
\hline
\end{tabular}
}
\end{center} 
\caption{\label{t:deptag-cpm} BLEU-4 scores (\%) of $tag$CNN model with dependency head words as additional tags ($tag$CNN\_dep).}
\end{table}

\begin{table}[t!]
\begin{center}
\scalebox{0.9}{
\begin{tabular}{l|ccc}
\hline
\textbf{Systems} & \textbf{MT04} & \textbf{MT05} & \textbf{Average} \\
\hline\hline
Dep2Str & 34.89      & 32.24         & 33.57 \\
\hline
+$in$CNN	            & 36.92    & 33.72    & 35.32 \\
\hline
+$in$CNN-$2$-pooling  & 36.33    & 32.88    & 34.61 \\
+$in$CNN-$4$-pooling  & 36.46    & 33.01    & 34.74 \\
+$in$CNN-$8$-pooling  & 36.57    & 33.39    & 34.98 \\
\hline
\end{tabular}
}
\end{center} 
\caption{\label{t:gate-pool} BLEU-4 scores (\%) of $in$CNN models implemented with gating strategy and $k$ max-pooling, where $k$ is of \{2, 4, 8\}.}
\end{table}

\subsection{Gating Vs. Max-pooling} \label{s:gate-pool}
In this section, we investigate to what extent that our gating strategy can improve the translation performance over max pooling, with the comparisons on $in$CNN model as a case study. For implementation of $in$CNN with max-pooling, we replace the local-gating (Layer-2) with max-pooling with size 2 (2-pooling for short), and global gating (Layer-4) with $k$ max-pooling (``$k$-pooling"), where $k$ is of $\{2,4,8\}$. Then, we use the mean of the outputs of $k$-pooling as the final input of Layer-5. In doing so, we can guarantee the input dimension of Layer-5 is the same as the architecture with gating. From Table~\ref{t:gate-pool}, we can clearly see that our gating strategy can improve translation performance over max-pooling by 0.34$\sim$0.71 BLEU points. Moreover, we find 8-pooling yields performance better than 2-pooling. We conjecture that this is because the useful relevant parts for translation are mainly concentrated on a few words of the source sentence, which can be better extracted with a larger pool size.

\section{Related Work} \label{s:related}
The seminal work of neural network language model (NNLM) can be traced to~\newcite{Bengio03aneural} on monolingual text. It is recently extended  by~\newcite{devlin2014} to include additional source context (11 source words) in modeling the target sentence, which is clearly most related to our work, with however two important differences: 1) instead of the ad hoc way of selecting a context window in~\cite{devlin2014}, our model covers the entire source sentence and automatically distill the context relevant for target modeling; 2) our convolutional architecture can effectively leverage guiding signals of vastly different forms and nature from the target.

Prior to our model there is also work on representing source sentences with neural networks, including RNN~\cite{cho2014,googleS2S} and CNN~\cite{kalchbrenner2013}. These work typically aim to map the entire sentence to a vector, which will be used later by RNN/LSTM-based decoder to generate the target sentence. As demonstrated in Section~\ref{s:expts}, the representation learnt this way cannot pinpoint the relevant parts of the source sentences (e.g., words or phrases level) and therefore is inferior to be directly integrated into traditional SMT decoders.

Our model, especially $in$CNN, is inspired by is the automatic alignment model proposed in~\cite{bahdanau2014neural}. As the first effort to apply attention model to machine translation, it sends the state of a decoding RNN as attentional signal to the source end to obtain a weighted sum of embedding of source words as the summary of relevant context. In contrast, $in$CNN uses 1) a different attention signal extracted from proceeding words in partial translations, and 2) more importantly, a convolutional architecture and therefore a highly nonlinear way to retrieve and  summarize the relevant information in source.

\section{Conclusion and Future Work} \label{s:conclusion}
We proposed convolutional architectures for obtaining a guided representation of the entire source sentence, which can be used to augment the $n$-gram target language model. With different guiding signals from target side, we devise $tag$CNN and $in$CNN, both of which are tested in enhancing a dependency-to-string SMT with +2.0 BLEU points over baseline and +1.08 BLEU points over the state-of-the-art in~\cite{devlin2014}. For future work, we will consider encoding more complex linguistic structures to further enhance the joint model.

\bibliographystyle{acl}
\bibliography{acl2015}
\end{document}

%% file: packages.tex
\usepackage{amsfonts, amssymb, amsmath}
\usepackage{algorithm, algorithmic, amsthm}
\usepackage{graphicx}
\usepackage{color}
\usepackage{hyperref}
\usepackage[top=2.5cm, bottom=3cm, left=3cm, right=3cm]{geometry}

\usepackage{subfigure}

%% file: MACPdef.tex
\newcommand{\w}{\ensuremath{\mathbf{w}}}
\newcommand{\x}{\ensuremath{\mathbf{x}}}

\newcommand{\z}{\ensuremath{\mathbf{z}}}

{%
\begin{list}{#1}{
\vspace{-\topsep}
\vspace{-\partopsep}
\setlength{\itemindent}{0cm}
\setlength{\rightmargin}{0cm}
\setlength{\listparindent}{0cm}
\settowidth{\labelwidth}{#1}
\setlength{\leftmargin}{\labelwidth}
\addtolength{\leftmargin}{\labelsep}
\setlength{\itemsep}{0cm}
}%
}%
{%
\end{list}
\vspace{-\topsep}
\vspace{-\partopsep}
}

{\begin{enumerate}%
}%
{\end{enumerate}}



%% file: MACPdef-ams.tex
\theoremstyle{plain}

\newtheorem*{lemma*}{Lemma}

\newtheorem*{prop*}{Proposition}

\theoremstyle{definition}

\newtheorem*{defn*}{Definition}

\newtheorem*{exmp*}{Example}

\newtheorem*{conj*}{Conjecture}

\theoremstyle{remark}

\newtheorem*{rmk*}{Remark}
